\pdfoutput=1

\documentclass[11pt]{article}

\usepackage{EMNLP2023}

\usepackage{times}
\usepackage{latexsym}
\usepackage{multirow}
\usepackage{rotating}
\usepackage{booktabs}
\usepackage[T1]{fontenc}

\usepackage[utf8]{inputenc}
\usepackage{xurl}

\usepackage{microtype}

\usepackage{inconsolata}

\title{Safe Training with Sensitive In-domain Data: Leveraging Data Fragmentation To Mitigate Linkage Attacks}

\author{Mariia Ignashina \\
  QMUL\\
  \texttt{m.ignashina@qmul.ac.uk} \\\And
  Julia Ive \\
  QMUL \\
  \texttt{j.ive@qmul.ac.uk} \\}

\begin{document}
\maketitle
\begin{abstract}
Current text generation models are trained using real data which can potentially contain sensitive information, such as confidential patient information and the like. Under certain conditions output of the training data which they have memorised can be triggered, exposing sensitive data. To mitigate against this risk we propose a safer alternative which sees fragmented data in the form of domain-specific short phrases randomly grouped together shared instead of full texts. Thus, text fragments that could re-identify an individual cannot be reproduced by the model in one sequence, giving significant protection against linkage attacks.
We fine-tune several state-of-the-art LLMs using meaningful syntactic chunks to explore their utility. In particular, we fine-tune BERT-based models to predict two cardiovascular diagnoses. Our results demonstrate the capacity of LLMs to benefit from the pre-trained knowledge and deliver classification results when fine-tuned with fragmented data comparable to fine-tuning with full training data.

\end{abstract}

\section{Introduction}
Healthcare and Social Sciences are increasingly turning to AI and Natural Language Processing (NLP) to create solutions for monitoring behavioral patterns and improving personalised treatments through textual analysis. In spite of recent progress, the full potential of NLP in those domains remains unexplored due to privacy concerns around data sharing.\footnote{\url{https://www.theguardian.com/society/2023/may/27/nhs-data-breach-trusts-shared-patient-details-with-facebook-meta-without-consent}} The issue has become more pressing due to the tendency to utilise cloud-based solutions for data storage and outsourcing of AI development to market leaders (such as OpenAI \footnote{\url{https://openai.com}}). 

 The intrinsic difficulty of anonymising textual data leads to public concern that these data may contain information that can directly or indirectly identify an individual. Such text can contain significant personal information through direct identifiers (names, IDs, etc.) and indirect identifiers (attributes that once combined with other knowledge, may lead to re-identification). 
 
 The scope of direct identifiers is well-defined and they are traditionally anonymised with Named Entity Recognition (NER) techniques \cite{tab23}. In contrast, the scope of indirect identifiers is unbound: any combination of textual chunks can potentially identify an individual causing a risk of a linkage attack \cite{lison-etal-2021-anonymisation}. Generalisable privacy-preserving ML approaches such as DP~\cite{Igamberdiev2023} are either not suitable for the anonymisation of text and degrade its quality, either too costly to deploy as in the case of Federated Learning~\cite{Yang2019}.

In this work we investigate sharing fragmented data. Such text chunks carry sense-bearing information, contain significantly fewer direct identifiers and are grouped together randomly to prevent linkage attacks. Our \textbf{main contribution} is hence a methodology to create training data composed of meaningful syntactic chunks to ease text sharing in privacy-sensitive scenarios. We benchmark our approach against the state-of-art of individual privacy protection (DP-Rewrite \cite{Igamberdiev2933}) and demonstrate the utility of our fragmented data to fine-tune LLMs to such downstream tasks as text generation and classification in the clinical domain. Our approach is generalisable, easy to deploy and opens attractive perspectives in terms of individual privacy protection.

\section{Related work}

Currently there is no reliable approach that handles privacy protection of textual data and considers their complex compositional nature. 

Traditionally, there are lines of approaches which only handle the protection of direct pre-set identifiers (such as name, address or insurance number) \cite{Hartman2020} or indirect pre-set identifiers (attributes that can identify a person in combination with other knowledge, e.g., such as age, gender or profession) without considering a broader scope of indirect identifiers (a combination of any textual chunks, rather than from a pre-defined set may breach privacy) \cite{olstad-etal-2023-generation}. Once detected, identifiers are removed (anonymisation), replaced with artificial identifiers (pseudoanonymisation) or with generalisations (sanitisation).

Recently, generalisable text anonymisation approaches have ported the most promising privacy-preserving Machine Learning (ML) techniques into Natural Language Processing (NLP) models. For example, the techniques of Differential Privacy (DP) re-write original text after injecting noise into meaning representations so that the resulting output can not be matched back to the original text \cite{Igamberdiev2023}. Only a few works investigate the capacity of such models to generate privacy-safe text (guarantee of individual privacy protection) \cite{Krishna2021,Igamberdiev2933,Igamberdiev2023} and so far these have been disruptive to the quality and utility of the outputs when sufficient privacy guarantees are respected. 

Another general ML method for privacy-safe adaptation is Federated Learning~\cite{Yang2019}. It enables learning, without data sharing, but performing noisy updates to central model weights from local models trained on confidential data. Those methods, however, require an infrastructure in place that enables multi-server communication which may be difficult to maintain in practice.    

In our work we propose a low-cost generalisable approach which offers attractive perspectives in terms of individual privacy preservation. So far to the best of our knowledge, only the utility of non-linguistic text chunks (phrases from Phrase-Based Statistical MT) has been successfully explored for domain adaptation in Machine Translation~\cite{kim-etal-2021-using}. Overall, releasing fragmented data for subsequent statistical analysis
has a long tradition in NLP (e.g., Google N-grams~\cite{Michel2011}).

\section{Methodology}

In our approach we extract noun phrases (NP) and verb phrases (VP) from our data using the Stanza \cite{qi2020stanza} toolkit. We consider nested constituents of length $2 \geq l \leq 4$. This allows us to obtain a sufficient volume of training data. Choosing meaningful syntactic units allows for more efficient selection of sense-bearing information, as well as more control over confidential information in the shared data (e.g., rare constituents can be filtered out).   

Following on from this we mix together those constituents to make sure that new training examples are not formed by fragments coming from the same original example. Such a way of creating the data drastically reduces the chances of linkage attack to succeed.

We do not use individual constituents as training examples since this approach may bias the downstream models, especially in the case of language modelling. To form each training example, we concatenate two NPs and two VPs.


\section{Experimental setup}

\subsection{Data} 

In our experiments we used sensitive medical data from MIMIC-III \cite{Johnson2016}, a publicly available ICU database. MIMIC-III contains anonymised records of ~40K patients admitted to a critical care unit of the Beth Israel Deaconess Medical Center between 2001 and 2012. The database is representative for cardiovascular diseases. We follow the best practices in the domain and focus on phenotyping from discharge summaries\footnote{In the medical text, the word ``phenotype'' refers to deviations from normal morphology, physiology, or behaviour, such as skin rash, hypoxemia, neoplasm, etc.} \cite{Harutyunyan2019}.

The initial dataset contained 25 phenotypes. We chose to work with the two most frequent ones: blood pressure (BP) and heart attack (HA). The total amount of notes which we extracted from MIMIC is 12K and 20K, respectively. Both prediction tasks are sparse classification with only 10\% of examples belonging to the cases and 90\% of controls. 10\% of each task's data was finally chosen for testing purposes (1.5K for BP and 2.2K for HA).

We then extracted syntactic chunks from the training part. The size of the fragmented training data is 23K and 42K, for BP and HA respectively. Our fragmented training examples are 24 tokens long on average, with a maximum of 32 tokens. We made sure to respect 10-90 split while creating the fragmented corpus.

MIMIC-III is anonymised for a pre-defined set of Protected Health Information (PHI) identifiers. We investigated how many times some of those identifiers (names, locations, drug names and occupations) appear in the fragmented data. We chose those identifiers that are more challenging to match using existing rule-based procedures.

Results in Table~\ref{Tab:ident} show that our data fragmentation procedure has the potential to reduce by half the percentage of such identifiers, with the most pronounced impact being on the quantity of location identifiers which were decreased by a factor of 7.

\begin{table}[h!]
   \centering
   \footnotesize
   \begin{tabular}{l|cc}
        \hline\hline
        \rule{0pt}{1ex} 
        Identifier & Full, \% of words & Frag \% of words \\
        \hline
Name &  0.45 & 0.37 \\
Location & 0.60 & 0.09 \\
Occupation  &  0.0001 &  0.00 \\
Drug  &  0.0004 & 0.00 \\
\hline
All  & 1 & 0.40 \\
        \hline\hline
    \end{tabular}
  \captionof{table}{Percentages of identifiers from total words in full and fragmented training data}
  \label{Tab:ident}
\end{table}

\paragraph{Data sharing scenario} In this study, we investigate a common scenario involving an external AI provider. The objective for the provider is to fine-tune their pre-trained models using client data. This provider is not trusted or only partially trusted; hence only fragmented data could be shared. Our goal is to assess whether these fragmented data will be useful for fine-tuning downstream task models, such as models for mainstream binary prediction and language modelling. We hence fine-tune relevant state-of-the-art models with full, fragmented and re-written texts. We test all our models on the same original test set. 

\subsection{Baseline} 

For our baseline we use the DP-Rewrite auto-encoder approach for rewriting original text to preserve its meaning with individual privacy guarantees. This is done by injecting Laplace noise into original meaning representations prior to rewriting. 

We employ DP-Rewrite with default parameters to rewrite  our data.\footnote{\url{https://github.com/trusthlt/dp-rewrite}} The rewritten training data are used to fine-tune downstream task models.

\subsection{Downstream tasks}

We use our privacy-safe data to fine-tune two mainstream models for the popular NLP tasks of language modelling and binary prediction.

For language modelling, we fine-tune the GPT-2 \cite{gpt} \texttt{small} model from the HuggingFace
Library \cite{wolf-etal-2020-transformers}. And we use the AdamW optimiser~\cite{adaw}. We then train each model with batch size of 3 to minimise
the cross-entropy loss over 3 epochs. For inference, we use the standard greedy sampling.

For binary prediction, we fine-tune  the BERT model from the
HuggingFace Library (\texttt{bert-base-uncased}). For each diagnosis the model was fine-tuned for 10 epochs using the Adam optimiser \cite{Kingma2014} with a learning rate of 3e-5.

\section{Results}

In this section, we present the results of our experiments using different privacy-preserving training data in the downstream tasks of language modelling and binary prediction.

\begin{table}[h!]
   \centering
   \footnotesize
   \begin{tabular}{l|cc}
        \hline\hline
        \rule{0pt}{1ex} 
        Data & Av. Acc. & Av. Proba \\
        \hline
Full &  0.46 & 0.45 \\
\hline
Frag & \bf 0.12 & \bf 0.12 \\
DP  &  0.05 &  0.04 \\
        \hline\hline
    \end{tabular}
  \captionof{table}{LM tasks result: average accuracy of the predicted word and average probability of the golden truth word (one word per sentence in the test set)}
  \label{Tab:lm-next-word}
\end{table}

\begin{table}[h!]
   \centering
   \footnotesize
   \begin{tabular}{p{20mm}|p{50mm}}
        \hline
        GPT FT Frag &  confused. denies pain. mostly had cxr. \\
        \hline
        GPT FT DP &  cardiovascular selling organs. cardiovascular selling organs.\\
                \hline\hline
        GPT FT Full & The pt id 65 y.o. female with good medical history. \\
\hline
    \end{tabular}
  \captionof{table}{Examples of clinical text generated by GPT fine-tuned with full texts, text fragments and text re-written by the baseline DP method (originally generated text has been paraphrased)}
  \label{Tab:lm-examples}
\end{table}

\paragraph{Language Modelling}

To assess the quality of the LM fine-tuning procedure, we analyse the performance of our models in the next word prediction task. For each sentence in our test data, we take the first five words as the prompt and ask our models to predict the next word. We measure the average accuracy of this prediction wrt golden truth words. We also report the average softmax probability of the golden truth word. Results are reported in Table~\ref{Tab:lm-next-word}. The first observation is that our fragmented data enable the LM model to outperform the fine-tuning with DP data by factor of two, for the latter the LM probabilities drop below 0.1. This demonstrates that our approach maintains data utility. Whilst there is a performance drop by factor of four when compared to the model fine-tuned with full data, we believe that this difference could be drastically reduced by fine-tuning an in-domain LLM (such as ClinicalGPT \cite{Wang2023}).

Table~\ref{Tab:lm-examples} presents some sampled text examples from our fine-tuned GPT models. GPT fine-tuned with fragmented data does produce slightly abrupt sentences with quite good adequacy, while the DP outputs are not coherent.

\paragraph{Diagnosis prediction}

To assess the quality of the BERT fine-tuning procedure, we analyse the performance of our classifiers in terms of precision, recall and F1-score.

\begin{table}[h!]
   \centering
   \footnotesize
   \begin{tabular}{l|ccc|ccc}
        \hline\hline
        \rule{0pt}{1ex} 
        & \multicolumn{3}{c}{BP} & \multicolumn{3}{c}{HA} \\
        Data & PR & RC & F1 & PR & RC & F1 \\
        \hline
 Frag & 0.49 & 0.53 & \bf 0.51 & 0.69 & 0.65 & \bf 0.67 \\
 DP & 0.40 & 0.46 & 0.42 & 0.63 & 0.58 & 0.60 \\
 \midrule
  Full &  0.55 & 0.52 & 0.53 & 0.72 & 0.73 & 0.72 \\
        \hline\hline
    \end{tabular}
  \captionof{table}{Results for the diagnoses prediction tasks: BP (blood pressure) and HA (heart attack). We report precision (PR), recall (RC) and F1-score (F1). Our models are trained using Full, Fragment and data re-written by the DP model. Results are averaged over 3 runs.}
  \label{Tab:results-class}
\end{table}

Table \ref{Tab:results-class} provides performance results across tasks and setups. As a general trend, the performance for HA (F1=0.72) is better than for BP (F1=0.53), which is not surprising given that there are twice more data for HA than for BP. This performance difference of almost 0.2 F1 score is maintained across setups with different privacy-preserving training data.

Models fine-tuned with our approach demonstrate an average improvement of +0.1 F1 as compared to DP. This confirms the utility of our data.

The loss of only around -0.04 F1 is observed on average across tasks as compared to using full texts. For BP, this loss is attributable to the reduction in precision (-0.06 PR) (precision contributes more to the performance in this data sparsity condition). For HA, it is attributable to the loss in recall (-0.08 RC) as there are more training data for this task.

\section{Conclusions}

With the increasing role of AI in the society, there is a growing need to share confidential text, which raises concerns regarding privacy. Current approaches to privacy-safe text sharing are either costly or suboptimal and detrimental for text utility. We propose a simple approach where fragmented data, consisting of short phrases specific to the domain, are shared instead of full texts. The resulting fragmented text has strong potential to preserve individual privacy since fragments combined together in new training examples do not come from the same original examples. 

By investigating the performance of state-of-the-art fine-tuned LLMs with these fragmented data, we demonstrate that these data remain useful for AI. This is evidenced by prediction models in the clinical domain that exhibit comparable performance when fine-tuned with fragmented and full texts. 

In this study we have explored the value of fragmented data for privacy-safe domain adaptation. We leave the investigation of relevant privacy guarantees and clinical validity to future work.
 

\section*{Ethics Statement}
 
 The proposed method of sharing fragmented texts useful for downstream AI analysis supports the protection of the fundamental human right to privacy. We envisage this will encourage increased ethical data sharing. Hence, we believe that the benefit of creating this technology outweighs the risks of deploying it. 

Before deployment in an real-life setting our methodology will need to pass rigorous benchmarking with privacy preservation metrics, as well as testing for clinical validity under the supervision of expert clinicians.
 
 The purpose of the clinical models presented in the paper is purely to demonstrate the utility of fragmented data. In the real-life scenario, downstream models employing our methodology will need to follow the ethical standards in the AI community with respect to privacy, fairness and transparency.

 The study has been carried out in accordance with relevant guidelines and regulations for the MIMIC database.

\section*{Limitations}

It is important to note that the methodology presented in this paper does not \textit{guarantee} individual privacy but is rather part of ongoing efforts towards designing shareable textual data. The key limitation of this work is the absence of the statistical analysis of the individual privacy guarantees. For this, we envisage using measures based on $k$-anonymity \cite{Sweeney2002} and exposure metrics for unintended memorisation \cite{Carlini-secret}. 

In the case of $k$-anonymity, we need to estimate whether some of the fragmented examples will have increased chances to be linked back to original training examples. As for the exposure metric, the chances of an identifier to appear in a syntactic chunk can be estimated.

Also, clinical validity of our fragmented text needs rigorous assessment by clinical experts. Combining syntactic chunks randomly can result in invalid or even erroneous clinical statements. Hence, additional restrictions on such combinations, as well as methods of their validation need to be established. 

Finally, the utility of chunked training data has been assessed only in a very limited number of scenarios. Further exploration is required to fully understand their advantages and disadvantages for a larger scope of downstream applications.  

 
\bibliography{anthology,custom}
\bibliographystyle{acl_natbib}

\appendix



\end{document}